\documentclass[11pt,a4paper]{article}
\usepackage[hyperref]{naaclhlt2019}
\usepackage{times}
\usepackage{latexsym}

\usepackage{graphicx}
\graphicspath{ {images/} }
\usepackage{subcaption}
\usepackage{bigstrut}
\usepackage{tikz,pgfplots}

\usepackage{multirow}
\usepackage{enumitem}
\usepackage{url}
\usepackage{amsmath}

\aclfinalcopy % Uncomment this line for the final submission
\title{Automatic learner summary assessment for reading comprehension}

\author{Menglin Xia \\
  Amazon Research Cambridge\thanks{The work by the first author was done at the University of Cambridge prior to joining Amazon Research.} \\
  {\tt ximengli@amazon.co.uk} \\\And
  Ekaterina Kochmar \\
  ALTA Institute\\
  Computer Laboratory \\
  University of Cambridge \\
  {\tt ek358@cl.cam.ac.uk} \\\And
  Ted Briscoe \\
  ALTA Institute\\
  Computer Laboratory \\
  University of Cambridge \\
  {\tt ejb@cl.cam.ac.uk}}

\date{}

\begin{document}
\maketitle
\begin{abstract}
 Automating the assessment of learner summaries provides a useful tool for assessing learner reading comprehension. We present a summarization task for evaluating non-native reading comprehension and propose three novel approaches to automatically assess the learner summaries. We evaluate our models on two datasets we created and show that our models outperform traditional approaches that rely on exact word match on this task. Our best model produces quality assessments close to professional examiners. 
\end{abstract}

\section{Introduction}

Summarization is a well-established method of measuring reading proficiency in traditional English as a second or other language (ESOL) assessment. It is considered an effective approach to test both cognitive and contextual dimensions of reading \citep{mc37}. However, due to the high time and cost demands of manual summary assessment, modern English exams usually replace the summarization task with multiple choice or short answer questions that are easier to score \citep{alderson2005assessing}. Automating the assessment of learner summarization skills provides an efficient evaluation method for the quality of the learner summary and can lead to effective educational applications to enhance reading comprehension tasks.

In this paper, we present a summarization task for evaluating non-native reading comprehension and propose three novel machine learning approaches to assessing learner summaries. First, we extract features to measure the content similarity between the reading passage and the summary. Secondly, we calculate a similarity matrix based on sentence-to-sentence similarity between the reading passage and the summary, and apply a Convolutional Neural Network (CNN) model to assess the summary quality using the similarity matrix. Thirdly, we build an end-to-end summarization assessment model using the Long Short Term Memory (LSTM) model. Finally, we combine the three approaches in a single system using a simple parallel ensemble modeling technique. We compiled two datasets to evaluate our models, and we release this data with the paper. We show that our models outperform traditional approaches that rely on exact word match on the task and that our best model produces quality assessments close to professional examiners.

\section{Related Work}

Most of the previous studies on summary assessment are aimed at evaluating automated summarization systems \cite{lin2004rouge, lin2003automatic, nenkova2007pyramid}. In contrast to this line of work, our goal is to assess human-written summaries rather than machine-generated ones. 

Within the educational domain, several applications have been developed to help students with their writing summarization skills. Summary Street \cite{wade2004summary} is an educational software designed for children to develop summarization skills. It asks students to write a summary to a reading passage, and scores the summary by using Latent Semantic Analysis (LSA) to construct semantic representations of the text. This system uses the cosine similarity score based on LSA as the sole indicator of content similarity. %??, which fails to capture the dependency of words in the text. 
OpenEssayist~\cite{Whitelock:2013} is an interactive system that provides students with the automated feedback about their summative essays. The system extracts the key words, phrases and sentences from the essays and helps the students to investigate their distribution in text and the potential implications for the clarity of the narrative. %This line of previous work mostly focuses on writing summary skills assessment.

The work that is most similar to ours is the automatic scoring of a summarization task by \citet{madnani2013automated}, who designed a task to measure the reading comprehension skills of American students. In their experiments, students were asked to write a four-sentence summary for each of the two three-paragraph reading passages, with the first sentence summarizing the whole passage and the following three sentences summarizing each paragraph. To build an automated system to score the summaries, they randomly select a student summary with the highest score as the reference, and use $8$ feature types to train a logistic regression classifier to predict the summary score. They train a separate classifier for each of the two passages, and obtain accuracy scores of $65\%$ and $52\%$ respectively, outperforming the most-frequent-score baselines of $51\%$ and $32\%$. Most of the features used in \citet{madnani2013automated} are based on verbatim overlap. Although such metrics prove to be effective in various tasks, they cannot capture the content similarity when paraphrasing or a higher level of abstraction are used in the summary.

Few studies have addressed summarization assessment at a higher level. More recently, \citet{Ruseti18} have used Recurrent Neural Networks (RNNs) to automatically score summaries. In their model, a concatenated representation of the summary and the text built from two separate RNNs as well as a complexity score of the text are fed to a fully connected layer to predict a real number between $[0, 1]$. This number is then mapped to $3$ discrete classes representing the quality of the summary using linear regression. Their best model achieves $55.24\%$ in accuracy on a dataset of $636$ summaries collected using Mechanical Turk.

In this paper, we address several limitations of previous work. We build a system that uses verbatim features as well as features capturing higher level of abstraction. First, we aim to build a generic system that can evaluate the quality of a summary without having to train a separate model for each text. Second, whereas \citet{madnani2013automated} use a student summary with the highest score as the reference to evaluate the candidate summary, our goal is to build a fully-automated system that does not require selecting a pre-defined reference. Third, we aim to explore features and structures capable of better modeling semantic similarity beyond verbatim overlap. 

\section{Data}

This section outlines the summarization task used in our experiments. First, learners, regardless of their proficiency level, were asked to read three reading passages extracted from the Cambridge English Exams dataset of \citet{xia2016text} at the lower (B1), upper intermediate (B2), and advanced (C1) levels of the Common European Framework of Reference for Languages (CEFR). Then they were asked to write a summary of $50$, $100$, and $120$ words for each of the three tasks.\footnote{The word limits on the summarization tasks are set to keep a relatively constant compression ratio between the summary and the length of the original passage.}

\subsection{Pilot study with simulated learner data}
\label{sec:sim-data}
Before launching the experiments with the actual language learners and in order to develop the automated summary evaluation system, we first ran a pilot study and collected ``simulated learner" summaries from $50$ members of our university. Since most participants of this study would speak English at an advanced (C1-C2) level, we asked them to write a ``good summary'' and a ``bad summary'' for each reading passage, mimicking a learner. The participants were asked to produce grammatically correct sentences and to write a ``bad" summary in a way that a learner who does not fully understand the original passage might produce. In total, $300$ summaries were collected (with $150$ good summaries and $150$ bad ones). The simulated learner data was then used to train binary classification systems to assess whether a summary captures the passage content properly or not.

\subsection{Real learner data}
\label{sec:summary_task}

\begin{table}[h]
	\centering
	%\resizebox{0.4\columnwidth}{!}{
	\begin{tabular}{|c|c|}
		\hline
		Learner levels & Count \\
		\hline
		B1 & 40 \\
		\hline
		B2 & 40 \\
		\hline
		C1-C2 & 57 \\
		\hline
		\hline
		Total & 137 \\
		\hline
	\end{tabular}
	%}
\caption{The distribution of the learner proficiency levels in the real learner data}
\label{dist_learner}
\end{table}

Next, we collected summaries from second language learners at B1, B2 and C1-C2 levels of proficiency.\footnote{The proficiency levels of learners were self-identified.} In total, $411$ summaries from $137$ learners were collected. The distribution of the learner proficiency levels is shown in Table \ref{dist_learner}.

\begin{figure}[ht]
	\centering
	\includegraphics[width=\columnwidth]{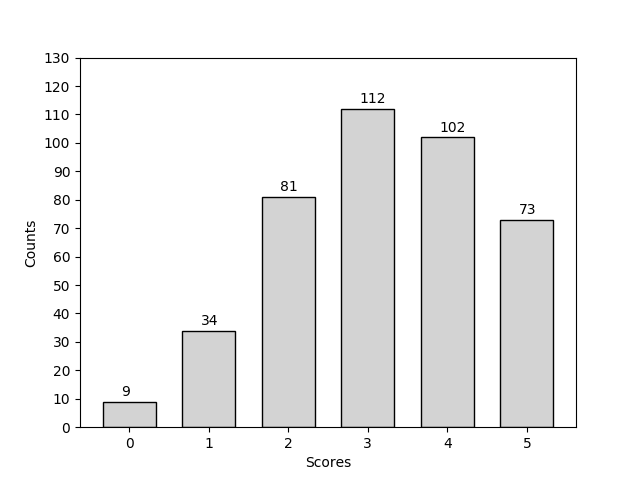}
	\caption{A histogram illustrating the score distribution in the real learner data}
	\label{img:dist-learner}
\end{figure}

The summaries were then scored by three professional ESOL annotators using a $6$-point scale defined as follows:

\begin{small}
\begin{itemize}[leftmargin=*]
    \item[] \textbf{Band 5:} \textit{The summary demonstrates excellent understanding of the original passage: 
	Content covers all the main points of the passage. 
	All content included is accurate, with no irrelevant details or repetitions. Target reader is fully informed.}

\item[] \textbf{Band 4:} \textit{Performance shares features of Bands 3 and 5.}
	
\item[] \textbf{Band 3:} \textit{The summary demonstrates acceptable understanding of the passage:
		Most of the main points are included.
		Most of the content is relevant and paraphrased, with some irrelevant details, repetitions or inaccuracy of content.  
		Target reader is on the whole informed.
	}

\item[] \textbf{Band 2:} \textit{Performance shares features of Bands 1 and 3.}

\item[] \textbf{Band 1:} \textit{The summary demonstrates very little understanding of the passage: 
		Most of the content is of limited relevance, with repetitions or verbatim borrowing from the original text. In some paraphrased parts of the text, inaccuracy of content or omissions of main points are evident. Target reader is minimally informed.
	}

\item[] \textbf{Band 0:} \textit{No understanding of the passage is demonstrated. The content is totally irrelevant to the original passage. Target reader is not informed.}
\end{itemize}
\end{small}

Figure \ref{img:dist-learner} shows the distribution of the scores for the learner summaries. The pairwise correlation between annotators ranges between $0.690$ and $0.794$. To derive the final score for each summary, we take the average of the scores by the three annotators. This results in a set of real-valued average scores on the scale of $[0, 5]$ and allows us to treat this task as a regression problem and make use of the continuity of the assessment scale. The goal of the experiments on this data is then to train a regression model to predict a score that correlates well with the annotators' judgments. 

\section{Methods}
In this section, we introduce three different approaches to the automated evaluation of the learner summaries.

\subsection{Measures for summary assessment}
First of all, we extract a number of features to describe the similarity of the summary and the reading text and apply a machine learning model to predict the summary quality.

The summarization task for reading comprehension examines the content relevance and the ability to convey the main ideas of the text in the summary. To automatically assess the learner summary, we compare the candidate summary against a reference to assess the quality of its content. 

We experiment with two types of references to evaluate the candidate summary: firstly, we compare the candidate summary against the \textit{original passage} directly, and secondly, we extract key sentences from the original text with an automated extractive summarizer and compare the candidate summary to the set of key sentences. Ideally, an extractive summarizer extracts a subset of sentences from the passage that are highly representative of the original text. Although the extracted key sentences are not necessarily coherent among themselves, they provide a representation of the main ideas of the text. Comparing the candidate summary against the key sentences allows us to examine the content relevance and the coverage of the main ideas in the candidate summary. We compare two popular summarizers in selecting the key sentences for reference: \textit{TextRank} \citep{mihalcea2004textrank} and \textit{MEAD} \citep{radev2004mead}. We also compare the extractive summarizers against the baseline of using a \textit{random selection} of sentences as the reference.

After obtaining the reference, we derive four types of linguistic features to evaluate the quality of the learner summary: (1) verbatim features, (2) semantic similarity features, (3) features based on distributed vector representations of the summary, and (4) features that describe discourse and other textual characteristics of the summary.

\subsubsection{Verbatim features}
Verbatim similarity is the most straightforward measure that indicates content similarity. Verbatim features measure the lexical overlap of the text units between the candidate summary and the reference. We use the following metrics to measure verbatim similarity: \textbf{ROUGE} \cite{lin2004rouge}, \textbf{BLEU} \cite{papineni2002bleu}, and \textbf{METEOR} \cite{denkowski2011meteor}. The three metrics are commonly used to assess automated summarization systems. ROUGE and BLEU are based on exact word match of N-grams, and METEOR extends the exact word match with stem, synonym, and paraphrase matches extracted from the WordNet \cite{miller1995wordnet} and a background dictionary, which allows for more flexible expressions.
	
\subsubsection{Semantic similarity features}
\label{sec:sentvec}
Although verbatim overlap metrics prove to be effective in various tasks, they fail to capture the content similarity when paraphrasing and higher levels of abstraction are used in the summary. To compensate for this, word embeddings and sentence embeddings are used to model semantic similarity at the word and the sentence level. We measure the semantic similarity between words and sentences in the texts and combine the scores into a measure of document-level semantic similarity. 

\begin{enumerate}
	\item \textbf{Word similarity}: \textbf{Word2vec} \cite{mikolov2013distributed} is a model for learning distributed vector representations of words from a large corpus of text. We use embeddings pre-trained on Wikipedia to compute word-to-word cosine similarity between the candidate summary and the reference. We experiment with three scoring functions to construct the text-level semantic similarity measures from the word-to-word scores:
	
	(1) \textit{average word similarity} on every word pair in the candidate summary and the reference; 
	
	(2) \textit{a greedy method} \cite{mihalcea2006corpus} that finds the best-matching word with maximum similarity scores and computes the average over the greedily selected pairs;

	(3) \textit{optimal matching} \cite{rus2012comparison} that finds the optimal alignment of word pairs and then takes the average over the alignment.
	 
	\item \textbf{Sentence similarity}: \textbf{Skip-thought} \cite{kiros2015skip} is a model for learning distributed representations of sentences. It uses an RNN-encoder to compose the sentence vector, and a decoder conditioned on the resulting vector that tries to predict the previous and the next sentences in the context. We use the model pre-trained on the BookCorpus \cite{zhu2015aligning} to generate our sentence vectors. Additionally, we experiment with composing the sentence vectors using word embedding summation and taking the average (\textbf{average word embeddings}). We use the same functions for word-level similarity to compute the text semantic similarity from the sentence vectors. 
\end{enumerate}

\subsubsection{Distributed vector representations of the summary}
In addition to the word and sentence similarities, we investigate methods to model the content similarity between the candidate summary and the reference directly at the document level.

Specifically, we use the following five approaches to construct vector representations of learner summaries:

\textbf{TF-IDF} is a common method to construct document representations in information retrieval. TF-IDF weighted document vectors are frequently used for measuring query-document similarity.
	
\textbf{Doc2Vec} \cite{le2014distributed} is a neural network model for learning distributed representation of documents. We use the ``distributed memory of paragraph vectors (PV-DM)" variant of the model to construct our vector representation of the summary. The PV-DM model maps the document to a vector space and uses a combination of the document vector and the vectors of surrounding words to predict a target word.
	
\textbf{Latent Semantic Analysis (LSA)} \cite{landauer2006latent} applies singular value decomposition (SVD) on the term-document matrix to obtain vector space representation of documents.

\textbf{Latent Dirichelet Allocation (LDA)} \cite{blei2003latent} represents the documents as mixtures of topics. It can be used to measure the content similarity and topical relevance of documents.
	
We also make use of the \textbf{average word embeddings} to encode the summary. 

We use the Simple English Wikipedia corpus\footnote{\url{https://simple.wikipedia.org}} as our background resource to learn the document representations. The Simple English Wikipedia data is used to train the models because its documents are rendered simple for English learners. Therefore, the lexical usage and syntactic structure in Simple English Wikipedia are more similar to the summaries written by learners. We take the cosine similarity between the candidate and the reference vectors to evaluate their similarity.
	
\subsubsection{Discourse and other textual features}

Apart from the content-based measures of the summary, the textual quality of the summary is also important for its overall quality estimation. For instance, good summaries tend to be more coherent and logically consistent. We extract \textbf{lexical chain}-based discourse measures to assess the coherence of the text. Lexical chains model the semantic relations among entities throughout the text. We implement the lexical chaining algorithm developed by \citet{galley2003improving} and extract $7$ lexical chain-based features.\footnote{Features include: number of lexical chains per document, number of lexical chains normalized by text length, average/maximum lexical chain length, average/maximum lexical chain span, and the number of long chains.}

We also measure the following superficial textual features:

\textbf{Length}: Number of words in the summary.

\textbf{Compression ratio}: The ratio of the number of words in the summary to the number of words in the reading passage.

\textbf{Type-token ratio}: The ratio of the number of unique words to the total number of words in the summary.

\textbf{Text readability}: The reading difficulty (the CEFR level) of the passage to be summarized.

After the features are extracted, we train a Support Vector Machine (SVM) \citep{cortes1995support} model for the classification task (Section  \ref{sec:sim-data}) and a Kernel Ridge Regression (KRR) \cite{Saunders:1998:RRL:645527.657464} model for the regression task (Section \ref{sec:summary_task}).

\subsection{Assessing summary using similarity matrix}

Secondly, we construct a sentence similarity matrix between the candidate summary and the original reading passage and apply a Convolutional Neural Network (CNN) model on the similarity matrix to predict the quality of the summary.

\citet{lemaire2005computational} proposed a computational cognitive model for assessing extractive summarization. In their experiments, they presented $278$ American school students with two reading passages and asked them to underline three to five sentences that they considered the most important in the texts. The underlined sentences were compared against the set of all the sentences from the original passage. They observed that the important sentences selected by the students are highly connected to the rest of the sentences in the text, where the connection is defined by the semantic similarity of the sentences. 

Based on their observations, we hypothesize that sentences in a good summary should have a well-distributed connection with as many sentences as possible in the original text, because a good summary is supposed to cover all the important information in the text. In contrast, sentences in a bad summary may fail to form a well-distributed connection with sentences in the original text. For example, if a bad summary only captures a few of the main points in the original text, then the sentences in such a summary would be connected only to the sentences where these points are mentioned in the original text, lacking the connections to the rest of the text. If a bad summary is generally irrelevant to the original text, sentences in such a summary would be minimally connected to most of the sentences in the original text. Beside these extreme cases on summary quality scale, summaries of intermediate quality may display patterns of connection to the original passage that share the characteristics of the good summary and the bad summary to various degrees. 

\begin{figure}[t]
	\centering
	\begin{subfigure}[b]{\columnwidth}
		\includegraphics[width=\columnwidth]{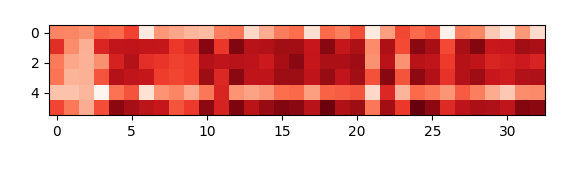}
		\caption{The similarity matrix of a good Summary A}
	\end{subfigure}
	\begin{subfigure}[b]{\columnwidth}
		\includegraphics[width=\columnwidth]{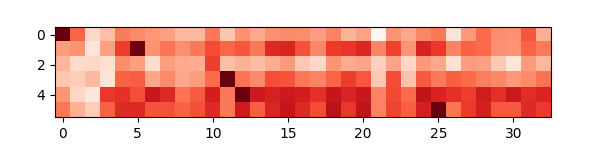}
		\caption{The similarity matrix of a bad Summary B}
	\end{subfigure}
	\caption{Similarity matrices of two summaries for the same reading passage from the simulated learner data. Summary A is a good summary and Summary B is a bad summary. The rows of the matrix represent sentences in the summary and the columns of the matrix represent sentences in the reading passage.}
	\label{fig:sim_mat}
\end{figure}

\begin{figure*}[t]
\centering
	\includegraphics[width=2\columnwidth]{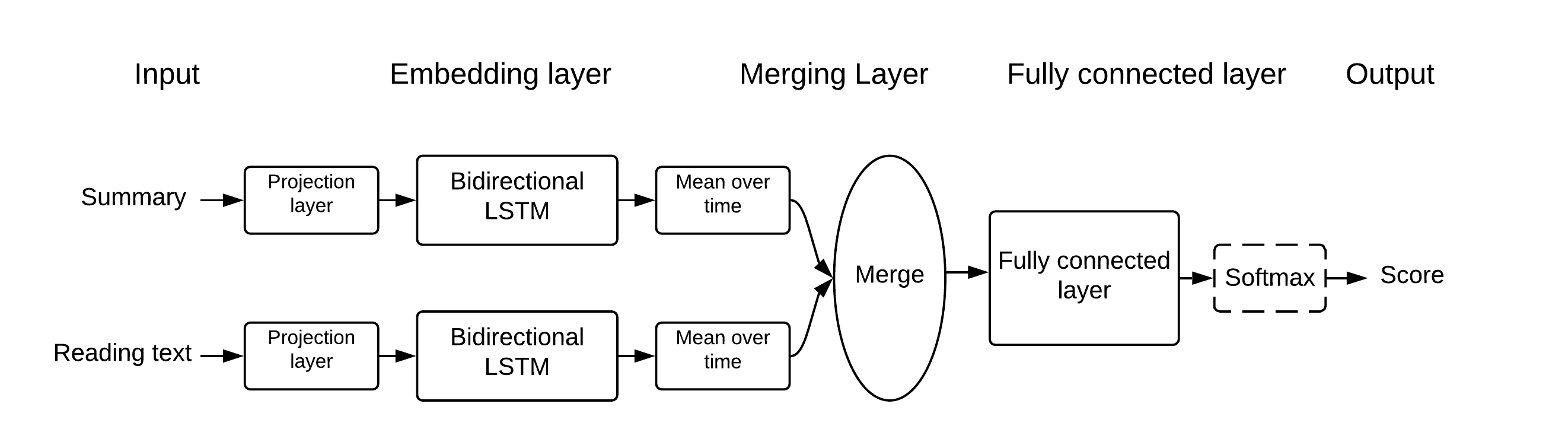}
	\caption{The merged LSTM model}
	\label{fig:merge_lstm}
\end{figure*}

Following this idea, we construct a sentence similarity matrix between the candidate summary and the original text. Each element of the matrix is a cosine similarity score between the vector representations of a sentence from the summary and a sentence from the original text. We use the two sentence similarity models described in Section \ref{sec:sentvec}, skip-thought and average word embeddings, to build the sentence vectors.

According to our hypothesis, the quality of the summary corresponds to different patterns in the similarity matrix. The similarity matrix can be viewed as a heat map ``image'' from which we can learn patterns to detect the quality of the summary. Figure \ref{fig:sim_mat} demonstrates the similarity matrices of two summaries for the same reading passage from the simulated learner data. The shade of the coloured map indicates the degree of similarity between two sentences: the darker the shade is, the more similar the sentences are. In this example, Summary A is an example of a good summary, and Summary B is an example of a bad summary. We can see that sentences in Summary A are similar to a number of sentences in the original text, resulting in a well-distributed heat map. By contrast, sentences in Summary B are similar to five particular sentences in the text and are less similar to other sentences, which is reflected by the isolated dark patches in the image. On the whole, Summary A has higher similarity scores than Summary B, which makes its heat map darker. These two examples illustrate how different patterns may be observed in the heat map of the summaries of different quality.

To learn these patterns automatically, we apply a CNN model on the similarity matrix to predict the quality of the summary. However, it should be noted that CNNs usually work best when a large amount of training data is available, whereas the summary data we have collected represents a relatively small dataset. We compare the results of the CNN model against the feature extraction approach to investigate how well the model can learn from the limited amount of data. 

\subsection{Assessing summary using LSTM-based models}

Thirdly, we experiment with several LSTM-based neural network models for assessing the summary quality. The LSTM-based models are used to learn representations of the summary and estimate its quality automatically, without having to manually extract features from it. 

Recurrent neural networks with LSTM units \cite{hochreiter1997long} have shown impressive results on various NLP tasks \cite{wang2016machine, rocktaschel2015reasoning}. In essence, they are capable of embedding long text sequences into a vector representation which can later be decoded for use in various applications.

\subsubsection{Merged LSTM model}

Inspired by the recent advances with LSTMs in NLP tasks, we propose a merged LSTM model (see Figure \ref{fig:merge_lstm}) for assessing learner summaries. The merged LSTM model encodes the summary and the reading text separately with two bidirectional LSTMs, and merges the embedded summary and embedded reading text representations into a joint representation to predict the summary score. We explore four functions to merge the encoded vectors, including {\em concatenation}, {\em addition}, {\em dot product} and {\em linear combination}.

%\begin{figure}[ht]
%	\includegraphics[width=1.1\columnwidth]{merge_lstm}
%	\caption{The merged LSTM model}
%	\label{fig:merge_lstm}
%\end{figure}

\subsubsection{Attention-based LSTM model}

As the merged LSTM model encodes the summary and reading text separately, it needs to propagate dependencies over long sequences to compare the summary and the text. The joint representation obtained in the merged LSTM model cannot fully capture the connection between the summary and the text. In this section, we propose an attention-based LSTM model which makes use of an attention mechanism to better model the relation between the summary and the reading text. 

The attention mechanism was first introduced by \citet{bahdanau2014neural} for machine translation. In general, the attention model learns a soft alignment between the input and the output in the encoder-decoder framework. The attention mechanism allows the model to learn what to attend to in the input states and mitigates the long-dependency bottleneck of the LSTM.

\begin{figure}[t]
	\centering
	\includegraphics[width=\columnwidth]{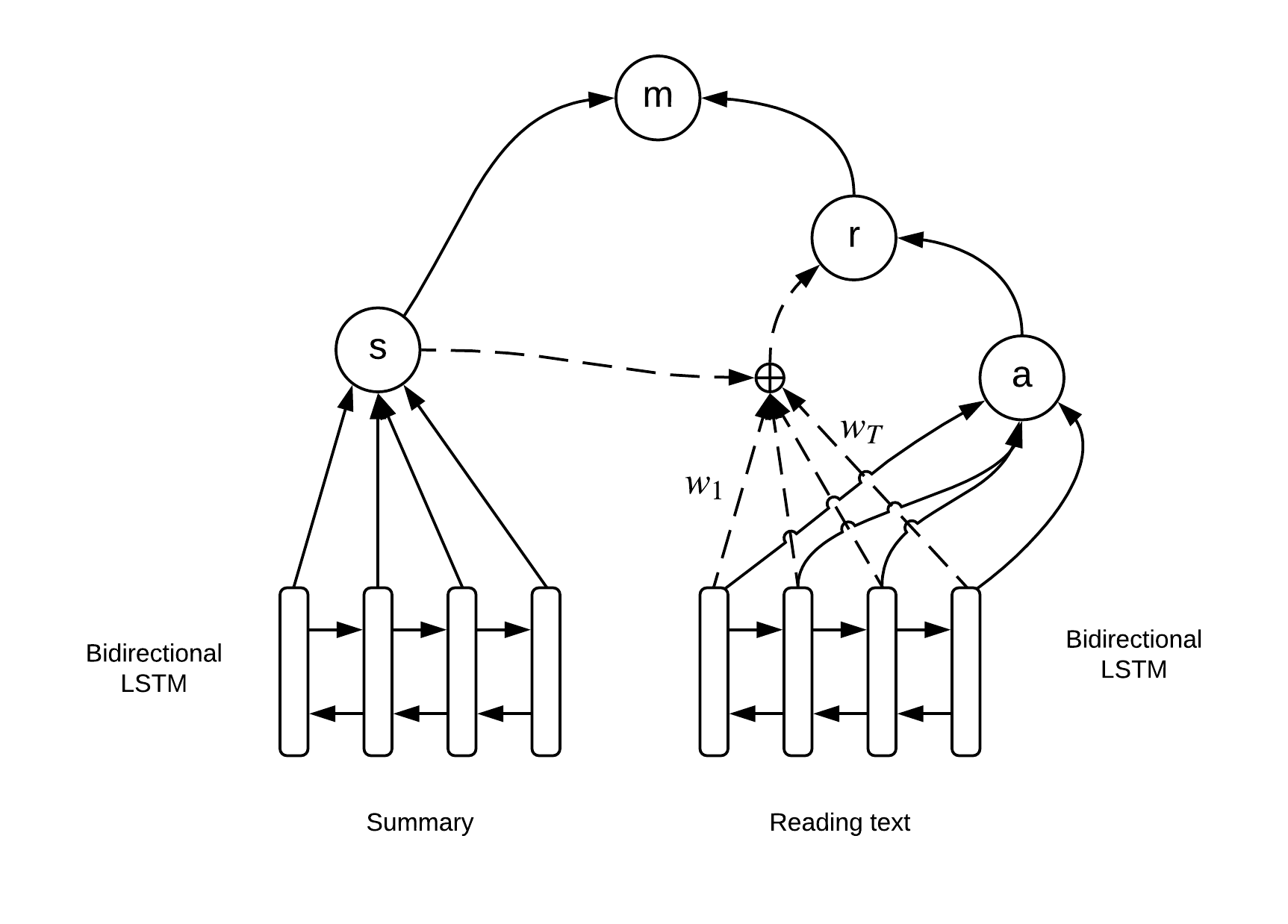}
	\caption{Attention mechanism architecture in the attention-based LSTM model for summary assessment}
	\label{fig:attention}
\end{figure}

In the attention-based model for summary assessment, the original text and the summary are still encoded using two separate LSTMs. However, the text representation is formed by a weighted sum of the hidden states of the text encoder, where the weights can be interpreted as the degree to which the summary attends to a particular token in the text. The summary representation and the text representation are combined with a nonlinear function into a joint representation and then fed into the fully connected layer to predict the summary quality. Figure \ref{fig:attention} is an illustration of the attention mechanism between the embedded summary and the embedded input text.

Mathematically, suppose $s$ is the encoded summary vector and $a(t)$ is the hidden state of the LSTM for the text at each token $t$. Then the final representation $r$ of the encoded text is a weighted sum of $a(t)$:

\[r = a \cdot w = \sum_{t=1}^{T} a(t)w(t)\]

The weight for each token $w(t)$ is computed by:

\[w(t) = \frac{\text{exp}(\alpha(t))}{\sum_{t=1}^{T} \text{exp}(\alpha(t))}\]

where \[ \alpha(t) = W_{a\alpha} \cdot a(t) + W_{s\alpha}\cdot s\]
is an alignment model.%, and $W_{a\alpha}$ and $W_{s\alpha}$ are learnable parameters. 

The joint representation $m$ of the summary and the text is a combination of the summary vector $s$ and the weighted input text vector $r$.
\[m = \text{tanh}(W_{sm} * s + W_{rm} * r + b)\]
where $W_{sm}$, $W_{rm}$ and $b$ are the parameters of a linear combination function.

\begin{table*}[t]
	\centering
	\resizebox{0.7\textwidth}{!}{
	\begin{tabular}{|c||ccc||c|}
		\hline
		Models & \multicolumn{3}{c||}{Variants} & Accuracy \bigstrut\\
		\hline
		\hline
		\multirow{5}[10]{*}{Baseline} & \multicolumn{2}{c|}{\multirow{5}[10]{*}{Baseline type}} & most-frequent & 50.0\% \bigstrut\\
		\cline{4-5}          & \multicolumn{2}{c|}{} & random  & 50.0\% \bigstrut\\
		\cline{4-5}          & \multicolumn{2}{c|}{} & ROUGE & 59.3\% \bigstrut\\
		\cline{4-5}          & \multicolumn{2}{c|}{} & BLEU  & 51.7\% \bigstrut\\
		\cline{4-5}          & \multicolumn{2}{c|}{} & ROUGE + BLEU & 59.6\% \bigstrut\\
		\hline
		\hline
		\multirow{4}[8]{*}{SVM} & \multicolumn{2}{c|}{\multirow{4}[8]{*}{reference type}} & random  & 58.8\% \bigstrut\\
		\cline{4-5}          & \multicolumn{2}{c|}{} & TextRank & 63.8\% \bigstrut\\
		\cline{4-5}          & \multicolumn{2}{c|}{} & MEAD  & 62.9\% \bigstrut\\
		\cline{4-5}          & \multicolumn{2}{c|}{} & original text  & \textbf{65.6\%} \bigstrut\\
		\hline
		\hline
		\multirow{2}[4]{*}{CNN} & \multicolumn{2}{c|}{\multirow{2}[4]{*}{similarity matrix type}} & avg word embeddings & \textbf{65.8\%} \bigstrut\\
		\cline{4-5}          & \multicolumn{2}{c|}{} & skip-thought vectors & 63.4\% \bigstrut\\
		\hline
		\hline
		\multirow{5}[10]{*}{LSTM} & \multicolumn{1}{c|}{\multirow{4}[8]{*}{Merged LSTM}} & \multicolumn{1}{c|}{\multirow{2}[4]{*}{merging }} & concatenation & 68.0\% \bigstrut\\
		\cline{4-5}          & \multicolumn{1}{c|}{} & \multicolumn{1}{c|}{} & addition & 68.1\% \bigstrut\\
		\cline{4-5}          & \multicolumn{1}{c|}{} & \multicolumn{1}{c|}{\multirow{2}[4]{*}{function}} & multiplication   & 69.1\% \bigstrut\\
		\cline{4-5}          & \multicolumn{1}{c|}{} & \multicolumn{1}{c|}{} & linear combination & 70.4\% \bigstrut\\
		\cline{2-5}          & \multicolumn{3}{c||}{Attention LSTM} & \textbf{71.1\%} \bigstrut\\
		\hline
		\hline
		Combined model & \multicolumn{3}{c||}{SVM+CNN+LSTM} & \textbf{75.3\%*} \bigstrut\\
		\hline
	\end{tabular}%
	}
	\caption{Model performance on the simulated learner data. We use the bold font to indicate the best model for each method. The asterisk sign indicates the best performance across all models.}
	\label{tab:sim_res}%
\end{table*}%

\subsection{Ensemble modelling}

\begin{figure}[t]
    \centering
    \includegraphics[width=\columnwidth]{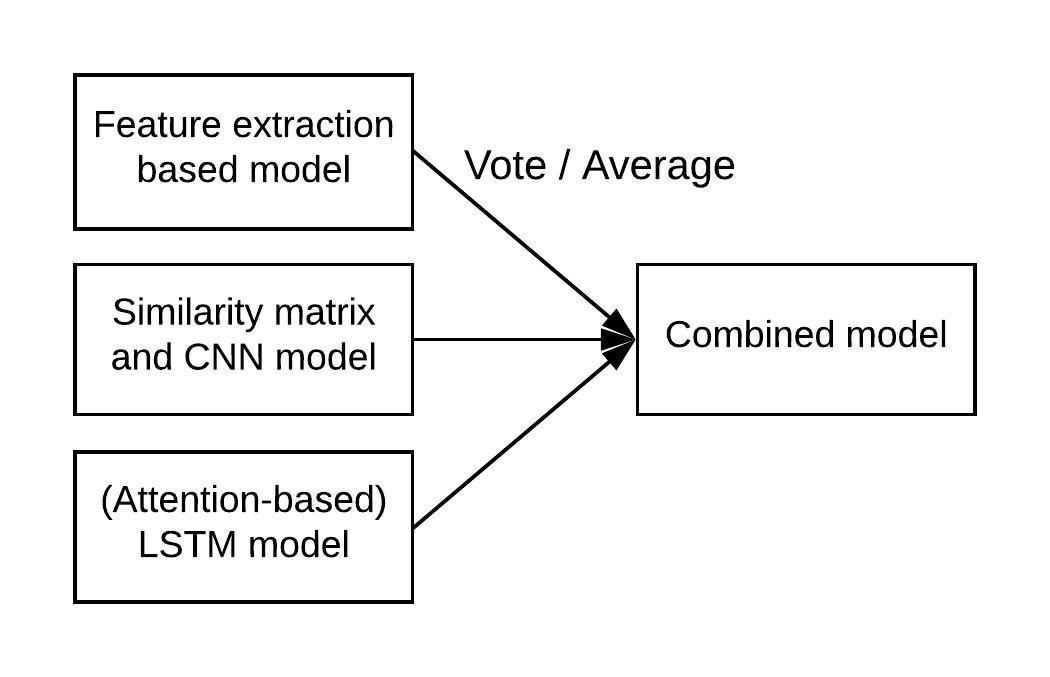}
    \caption{Combining three approaches using ensemble modelling}
    \label{fig:ens}
\end{figure}

Ensemble modelling combines several machine learning techniques into one model in order to improve the stability and accuracy of the prediction. We explore combining the three different models (see Figure \ref{fig:ens}) into a single model by taking the majority vote from the binary classification models and taking the average value of the predicted scores from the regression models. We compare the performance of the combined models against the individual models to investigate if and to what extent ensemble modelling is useful for assessing the summaries.

\section{Experiments and Results}

\subsection{Experimental set-up}

We evaluate our models on the real learner data and on the simulated learner data, for consistency, using 5-fold cross validation. In each fold, $60\%$ of the data is used as the training set, $20\%$ as the development set, and $20\%$ as the test set.\footnote{We choose the best model based on the development set, retrain the selected model on the combination of the training and the development data, and evaluate the model on the test set.}

We compare our models against five baselines: \textit{most frequent baseline}, \textit{random baseline}, \textit{ROUGE baseline},\footnote{A baseline trained on ROUGE features only.} \textit{BLEU baseline}, and \textit{ROUGE + BLEU baseline}.

We use accuracy to evaluate the models on the simulated learner data, and on the real learner data, we report scores of two evaluation metrics: Pearson correlation coefficient (PCC) and Root Mean Square Error (RMSE), which are commonly used for evaluating regression models.

\subsection{Results}

%Table \ref{tab:sim_res} and Table \ref{tab:learner_res} show the results of the baseline models and the results of the four types of models on the simulated learner data and the learner data respectively.

Table \ref{tab:sim_res} shows the results of the baseline and the four types of models on the simulated learner data, and Table \ref{tab:learner_res} reports the results of the models on the real learner data.

On the simulated learner data, the best variants from all three methods outperform the baselines. The improvement is statistically significant ($p$$<$$0.05$) using $t$-test for all three methods. We combine the best variants from the three approaches into a single system by taking the majority vote from the models. The resulting system achieves the best accuracy of $75.3\%$ in predicting the binary type of the summary on the simulated learner data.

% Table generated by Excel2LaTeX from sheet 'Sheet3'
\begin{table*}[t]
  \centering
  \resizebox{0.7\textwidth}{!}{
    \begin{tabular}{|c||ccc||c|c|}
    \hline
    Models & \multicolumn{3}{c||}{Variants} & PCC  & RMSE \bigstrut\\
    \hline
    \hline
    \multirow{5}[10]{*}{Baseline} & \multicolumn{2}{c|}{\multirow{5}[10]{*}{Baseline type}} & most-frequent & -     & 1.30  \bigstrut\\
\cline{4-6}          & \multicolumn{2}{c|}{} & random  & 0.011 & 1.79  \bigstrut\\
\cline{4-6}          & \multicolumn{2}{c|}{} & ROUGE & 0.499 & 1.12  \bigstrut\\
\cline{4-6}          & \multicolumn{2}{c|}{} & BLEU  & 0.208 & 2.88  \bigstrut\\
\cline{4-6}          & \multicolumn{2}{c|}{} & ROUGE + BLEU & \textbf{0.499} & \textbf{1.11}  \bigstrut\\
    \hline
    \hline
    \multirow{4}[8]{*}{KRR} & \multicolumn{2}{c|}{\multirow{4}[8]{*}{reference type}} & random  & 0.517 & 1.11  \bigstrut\\
\cline{4-6}          & \multicolumn{2}{c|}{} & TextRank & 0.576 & 1.06  \bigstrut\\
\cline{4-6}          & \multicolumn{2}{c|}{} & MEAD  & 0.557 & 1.08  \bigstrut\\
\cline{4-6}          & \multicolumn{2}{c|}{} & original text  & \textbf{0.636} & \textbf{0.99} \bigstrut\\
    \hline
    \hline
    \multirow{2}[3]{*}{CNN} & \multicolumn{2}{c|}{\multirow{2}[3]{*}{similarity matrix type}} & avg word embeddings & \textbf{0.504} & \textbf{1.12} \bigstrut\\
\cline{4-6}          & \multicolumn{2}{c|}{} & skip-thought vectors & 0.458 & 1.14  \bigstrut[t]\\
    \hline
\hline
    \multirow{5}[9]{*}{LSTM} & \multicolumn{1}{c|}{\multirow{4}[7]{*}{Merged LSTM}} & \multicolumn{1}{c|}{\multirow{2}[3]{*}{merging }} & concatenation & 0.487 & 1.13  \bigstrut[b]\\
\cline{4-6}          & \multicolumn{1}{c|}{} & \multicolumn{1}{c|}{} & addition & 0.466 & 1.13  \bigstrut\\
\cline{4-6}          & \multicolumn{1}{c|}{} & \multicolumn{1}{c|}{\multirow{2}[4]{*}{function}} & multiplication & 0.490 & 1.12  \bigstrut\\
\cline{4-6}          & \multicolumn{1}{c|}{} & \multicolumn{1}{c|}{} & linear combination & 0.484 & 1.13  \bigstrut\\
\cline{2-6}          & \multicolumn{3}{c||}{Attention LSTM} & \textbf{0.494} & \textbf{1.12}  \bigstrut\\
    \hline
    \hline
    Combined model & \multicolumn{3}{c||}{KRR+CNN+LSTM} & \textbf{0.665*} & \textbf{0.97*}  \bigstrut\\
    \hline
    \end{tabular}%
}
  \caption{Results of the regression model performance on the learner data. We use the bold font to indicate the best model for each method. The asterisk sign indicates the best performance across all models.}
  \label{tab:learner_res}%
\end{table*}%

On the real learner data, we found that the feature extraction-based model outperforms the CNN model and LSTM-based models, which also significantly outperform the baselines. The results suggest that the neural network-based models are not as effective as the traditional feature extraction-based method for the regression task, at least when the training data is limited in size.

However, although the CNN and LSTM models are not the best-performing models individually, a combination of the three methods (KRR, CNN and LSTM) still improves the performance. We believe that this is because the three independent models capture different aspects of the summary quality that are complementary to each other. In addition, the combined model is more robust to outliers. For example, when two models agree on an instance while the third model does not, the combined model will select the majority vote or the average score of the model estimations, hence achieving a better performance in estimating the summary quality. Overall, the best model performance is close to human performance.

We also observe that when assessing the summaries with extracted features, using the original document as the reference works better than using other types of reference. This might be because the extractive summarizers only select sentences that are highly related to others, where the relation is typically judged by the word overlap, therefore missing the bits of text where topical words occur less often. 

\section{Conclusion}

In this paper, we introduce a summarization task for testing reading comprehension of learners and present several automated systems to assess the quality of the learner summary. We collected summaries from members of our university and from the real learners to evaluate our systems. We propose and compare three approaches to assess the summaries, including the feature extraction-based model, the CNN-based model using similarity matrix, and the LSTM-based model. The best system, built using a combination of three models, yields an accuracy of $75.3\%$ on the simulated learner data, and $PCC=0.665$, $RMSE=0.97$ on the real learner data. Although not directly comparable to other studies, we note that these results are higher than those reported in previous work.

Our systems are generalizable and address the limitations of the previous research in this area as: (1) they are capable of evaluating the quality of a summary without the need of being trained on each input text separately, (2) they do not require a pre-defined reference, and (3) they are capable of capturing content similarity beyond verbatim overlap, taking into account paraphrasing and higher levels of abstraction.

We believe that although the application presented in this paper focuses on assessing learner summaries, these techniques may also be useful for benchmarking automated summarization systems. Evaluation of these techniques for benchmarking automated summarization systems is one direction for our future research.

We make the summary data available at \url{https://www.cl.cam.ac.uk/~ek358/learner-summaries.html}.

\section*{Acknowledgments}

% The acknowledgments should go immediately before the references.  Do
% not number the acknowledgments section. Do not include this section
% when submitting your paper for review. \\

This paper reports on research supported by Cambridge Assessment, University of Cambridge. We also thank Cambridge Assessment for their assistance in the collection of the real learner data. We are grateful to the anonymous reviewers for their valuable feedback.

\bibliography{naaclhlt2019}
\bibliographystyle{acl_natbib}

\end{document}